%% file: main.tex
\documentclass[letterpaper,10pt,conference]{ieeeconf}

\IEEEoverridecommandlockouts
\usepackage[utf8]{inputenc}
\usepackage[T1]{fontenc}
\usepackage{amsmath,amssymb}
\usepackage{booktabs}
\usepackage{cite}
\usepackage{graphicx}
\usepackage{microtype}
\usepackage{tikz}
\usetikzlibrary{arrows.meta,positioning}
\usepackage{flushend}
\usepackage{hyperref}

\newcommand{\R}{\mathbb{R}}
\newcommand{\clip}{\operatorname{clip}}

\title{\LARGE \bf
Observation-Constrained Joint-Space Viewpoint Optimization\\
for Robotic Inspection of Cylindrical Cavities
}

\author{Yuezhong Wang$^{*}$, Rongshen Yin$^{*}$, Bichi Zhang, and S\"oren Schwertfeger
\thanks{$^{*}$These authors contributed equally to this work.}
\thanks{Yuezhong Wang, Bichi Zhang, and S\"oren Schwertfeger are with the
Key Laboratory of Intelligent Perception and Human--Machine Collaboration,
Ministry of Education, ShanghaiTech University, Shanghai, China.
E-mail:
\texttt{\{wangyzh12024, zhangbch2025, soerensch\}@shanghaitech.edu.cn}.}%
\thanks{Rongshen Yin is with the University of Pennsylvania,
Philadelphia, PA, USA.
E-mail: \texttt{yinrsh@engineering.upenn.edu}.}%
\thanks{This work was supported by the National Natural Science Foundation of China under Grant W2531052. The experiments were supported by the Core Facility Platform of Computer Science and Communication, SIST, ShanghaiTech University.}%
}

\begin{document}

\maketitle
\thispagestyle{empty}
\pagestyle{empty}

\begin{abstract}
Inspection is a core capability in many mobile robotics applications, including industrial facility monitoring, infrastructure maintenance, agriculture, and search and rescue. Observing the bottom of a cylindrical cavity, as required by ASTM search-task benchmarks for response robots, presents a representative challenge: the robot must position its camera precisely while satisfying visibility, kinematic, and collision constraints. This paper presents a fully autonomous method for observation-constrained inspection of cylindrical cavities in robot joint space. Rather than prescribing a single Cartesian camera pose, the method represents the inspection objective as a set of valid viewing geometries, thereby avoiding the rejection of reachable viewpoints and configurations with poor joint-limit margins. An RGB perception front end estimates the opening center and directed cavity axis from semantic masks using arc-supported ellipse fitting together with body and side-generator cues. These estimates parameterize constraints on camera-axis alignment, lateral offset, and axial standoff. A multistart derivative-free search then optimizes robot joint configurations with lexicographic priority given to constraint satisfaction; feasible configurations are ranked according to motion economy, joint-limit margin, and view quality. The resulting candidates are evaluated by a collision-aware motion planner, and the executed camera pose is verified geometrically and using a ray-based estimate of bottom visibility. In Isaac Sim, the proposed method
successfully completes 92 of 100 target configurations and attains 91.65\% mean bottom
visibility among executed trials, compared with 76 of 100 and 84.3\% for a
multistart coordinate-search baseline. Tabletop and Unitree A2-mounted
experiments demonstrate the complete perception--planning--execution pipeline.
\end{abstract}

\input{sections/01_introduction}

\input{sections/02_related_work}
\input{sections/03_method}
\input{sections/04_experiments}
\input{sections/05_conclusion}

\bibliographystyle{IEEEtran}
\bibliography{ref}

\end{document}

%% file: sections/01_introduction.tex
\section{Introduction}
\label{sec:introduction}

The ASTM E2853/E2853M-22 standard \cite{ASTM_E2853_E2853M_22} defines search tasks for evaluating ground response robot capabilities and is used in the RoboCup Rescue Robot League\cite{sheh2011robocuprescue}. These tasks include inspecting confined openings and exemplify the precise camera placement required for robotic inspection.

Robotic inspection differs from conventional pick-and-place manipulation in that the task objective is primarily observational. For a cylindrical cavity, a useful eye-in-hand view must look through the opening, remain near the cavity centerline, and preserve sufficient standoff to expose the interior. These requirements define a continuum of acceptable camera configurations. These requirements are also constrained by robot reachability: a visually ideal pose may approach a joint limit, lack a collision-free trajectory, or be unreachable from the robot’s current configuration.

This set-valued structure matters because pose error alone is not a sufficient
measure of inspection utility. Camera roll can vary without changing whether
the cavity bottom is visible, while a small translational change near the rim
can occlude a large part of the interior. Conversely, a configuration that is
not closest to a nominal camera pose may provide a better executable view after
joint limits and collision-free connectivity are considered. The planning
objective should therefore preserve the allowable observation region until
robot feasibility has been evaluated.

\begin{figure}[t]
    \centering
    \includegraphics[width=\columnwidth]{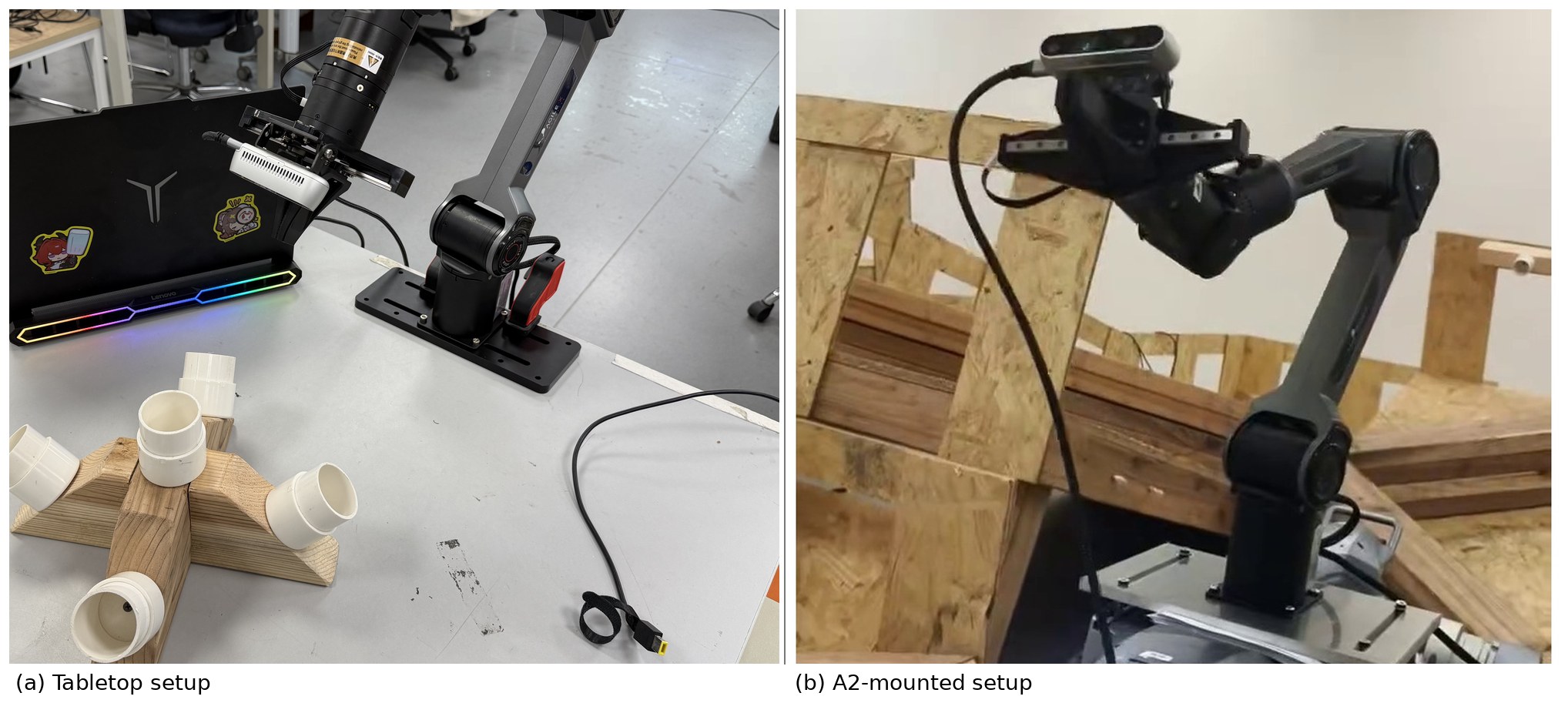}
    \caption{Inspection platforms used in this work: (a) the tabletop
    eye-in-hand setup and (b) the Piper manipulator mounted on a Unitree A2.\vspace{-.5cm}}
    \label{fig:system_overview}
\end{figure}

Most manipulation pipelines first choose a Cartesian target and then invoke an
inverse-kinematics (IK) solver. This decomposition is appropriate when the task
requires a unique end-effector pose, but it is unnecessarily restrictive for
inspection. View sampling relaxes the single-pose assumption, although it still
searches camera-pose space before evaluating manipulator feasibility. The visual
front end also faces target-specific projective effects: the fitted ellipse
center generally differs from the projection of the physical circle center
\cite{matsuoka2016eccentricity}, while unique single-view circle-based pose
recovery requires additional geometric information, such as a discriminable
circle center or a predefined object frame \cite{wang2008singleview}.

We formulate cylindrical inspection as a search over robot configurations
subject to explicit observation constraints. The visual front end provides only
the task-relevant opening center and a directed axis pointing into the cavity.
These quantities define an observation region in which the camera pose may
vary. A feasibility-first joint-space optimizer searches this region, retains
multiple candidate configurations, and delegates collision-free connection to
the motion planner. The actually achieved view is then checked rather than
accepting the nominal optimizer output. The evaluation separates two questions:
a controlled Isaac Sim study uses identical ground-truth target geometry to
compare viewpoint generation, whereas hardware demonstrations exercise the
complete RGB perception-to-execution pipeline.

The contributions of this paper are threefold:
\begin{itemize}
    \item an observation-region formulation for cylindrical inspection that
    replaces a prescribed 6-DoF camera pose with explicit axis-alignment,
    lateral-offset, and axial-standoff constraints;
    \item a feasibility-first joint-space optimizer that ranks multiple robot
    configurations using lexicographic hard and soft objectives, followed by
    collision-aware planning and post-execution geometric and visibility
    verification; and
    \item a task-specific RGB front end that estimates the opening center and
    directed interior axis, integrated with a controlled 100-configuration
    simulation comparison and tabletop and quadruped-mounted demonstrations.
\end{itemize}

%% file: sections/02_related_work.tex
\section{Related Work}
\label{sec:related_work}

\subsection{Cylindrical Perception}
A projected circular rim generally appears as an ellipse, but its image center
need not coincide with the projection of the physical circle center
\cite{matsuoka2016eccentricity}. Moreover, unique single-view circle-based pose
recovery requires additional information, such as a discriminable circle
center or a predefined object frame \cite{wang2008singleview}. Robust circle
and ellipse detection can exploit local arc support rather than indiscriminate
fitting to a complete, noisy contour
\cite{lu2017arcsupportcircle,lu2020arcsupportellipse}. These observations are
particularly relevant to inspection targets whose smooth surfaces provide weak
texture and whose opening boundary may be only partially supported by reliable
image gradients.

Our front end combines semantic localization with arc-supported fitting and
uses the visible cylindrical body to choose a directed interior axis. The output
is deliberately task-specific: the planner needs the opening center, inward
axis, cavity length, and aperture radius, but not an arbitrary object-frame roll
angle. This avoids introducing unobservable pose components into the subsequent
viewpoint objective.

\subsection{View Planning and Active Perception}
Classical view planning selects sensor poses for reconstruction or inspection
coverage \cite{scott2003viewplanning}. Recent systems extend this idea to online
next-best-view selection for mobile manipulators, joint viewpoint planning and
depth completion in clutter, and hand-eye active perception for optically
difficult objects \cite{naazare2022nbv,liu2025rasp,kennis2026heapgrasp}. Such
methods commonly optimize scene coverage, reconstruction uncertainty, or task
perception quality over a sequence of views. Our problem is narrower but more
geometrically explicit: a single inspection action must see through a finite
circular aperture and retain a prescribed fraction of the internal bottom.

Visual servoing instead closes the loop on image- or pose-space errors
\cite{hutchinson1996visualservo,chaumette2006visualservo}. The present method is
not a replacement for visual servo control. It generates a feasible goal set
before motion planning and verifies the executed view afterward; closed-loop
image regulation could be added between these two stages when online target
tracking is available.

\subsection{Constrained Manipulation and Viewpoint Generation}
Pose-set formulations such as Task Space Regions represent manipulation goals
as allowable Cartesian regions \cite{berenson2011tsr}, while analytical sensor
planning has characterized admissible views under resolution, focus, and
field-of-view constraints \cite{tarabanis1994visibility}. Improved IK solvers
such as TRAC-IK increase robustness for a specified target pose
\cite{beeson2015tracik}. These formulations establish that task constraints need
not reduce to one rigid pose, but a pose-space region must still be mapped to
robot configurations and connected by a feasible trajectory.

Our method evaluates cylindrical observation constraints directly from forward
kinematics and searches in joint space. This permits joint-limit margin and
motion economy to influence viewpoint selection before trajectory generation.
The resulting ranked configurations are then checked by MoveIt
\cite{coleman2014moveit}. Thus, the contribution lies in the interface between a
task-specific observation set and manipulator feasibility, rather than in a new
collision-checking or trajectory-planning algorithm.

%% file: sections/03_method.tex
\section{Method}
\label{sec:method}

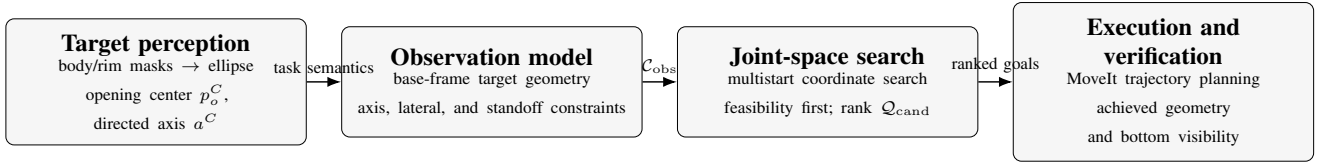
\begin{figure*}[t]
    \centering
    \resizebox{0.98\textwidth}{!}{\input{fig/method_pipeline.tex}}\vspace{-0.2cm}
    \caption{Pipeline overview. Perception estimates task-specific cylindrical
    geometry, which defines an observation region. Joint-space search returns
    multiple feasible configurations; motion planning and post-execution checks
    operate on the ranked set.\vspace{-0.5cm}}
    \label{fig:pipeline}
\end{figure*}

\subsection{Pipeline Overview}
\label{subsec:overview}

The framework exposes a compact semantic interface between perception and
manipulation. Perception returns an opening center $p_o^C$ and a unit axis
$a^C$ directed toward the cavity interior. Together with known target
dimensions, these quantities define a finite observation region rather than a
complete object pose. Forward kinematics maps each joint configuration to an
eye-in-hand camera pose, from which observation feasibility can be evaluated
without first constructing a Cartesian goal. The optimizer consequently returns
a ranked set of joint states. Motion planning tests this set in rank order, and
the final decision is made from the achieved robot state and estimated interior
visibility. Fig.~\ref{fig:pipeline} summarizes these stage interfaces.

\subsection{Problem Statement and Assumptions}
\label{subsec:problem}

Given an eye-in-hand image $I$ and current joint state $q_0$, the objective is
to return a ranked set of joint configurations whose forward-kinematic camera
poses satisfy the inspection geometry. The formulation assumes known opening
diameter $D$ and cavity length $L$, calibrated camera intrinsics, a fixed
eye-in-hand transform ${}^ET_C$, and a visible, static target. Target-free
hand--eye calibration with joint scene reconstruction provides one way to
obtain this transform \cite{zhi2017simultaneous}; Multical can additionally
register cameras, IMUs, and LiDARs spatially and temporally on a mobile platform
\cite{zhi2022multical}. Calibration is outside our scope, and all extrinsics are
fixed during inspection. The viewpoint optimizer
evaluates observation geometry and joint limits; collision geometry and
trajectory connectivity are delegated to MoveIt. This separation makes the
optimizer's output a set of candidate observation configurations rather than a
claimed collision-free motion.

\subsection{Geometry-Aware Target Perception}
\label{subsec:perception}

Let $I$ be the eye-in-hand RGB image. A hierarchical semantic front end first
localizes the cylinder body and then segments the opening inside the resulting
region of interest. To reduce sensitivity to reflections and mask leakage,
edge processing is restricted to a narrow band around the refined rim mask
$\hat M$,
\begin{equation}
\mathcal B_{\rm rim}=\operatorname{dilate}(\hat M,k_b)
-\operatorname{erode}(\hat M,k_b).
\label{eq:rim_band}
\end{equation}
Local arcs in $\mathcal B_{\rm rim}$ generate ellipse hypotheses. Candidates
are ranked by edge support, angular coverage, mask consistency, fitting
residual, and spatial distribution, after which the best inlier set is refitted
with direct least squares \cite{fitzgibbon1999ellipse}. Denote the selected
ellipse by $e=(u_e,v_e,A_e,B_e,\theta_e)$, where $A_e\geq B_e$ are the
projected major- and minor-axis diameters in pixels.

Given the calibrated focal length $f_x$ and known physical opening diameter
$D$, the implementation uses the approximate relations
\begin{equation}
Z_e=\frac{f_xD}{A_e},\qquad
\rho=\frac{B_e}{A_e},\qquad
\tau=\arccos(\rho),
\label{eq:ellipse_geometry}
\end{equation}
where $Z_e$ is the opening depth and $\tau$ is its inclination. Let
$d_m=[\cos(\theta_e+\pi/2),\sin(\theta_e+\pi/2)]^T$ be the image minor-axis
direction. Perspective eccentricity is handled by two bounded center
corrections,
\begin{equation}
\begin{aligned}
p_s&=[u_e,v_e]^T+s\Delta_c d_m,\\
\Delta_c&=\min\!\left(k_cA_e(1-\rho),k_{\max}B_e\right),
\end{aligned}
\qquad s\in\{-1,+1\}.
\label{eq:center_candidates_2d}
\end{equation}
The corresponding 3-D center and directed-axis hypotheses are
\begin{equation}
C_s^C=
\begin{bmatrix}
(u_s-c_x)Z_e/f_x\\
(v_s-c_y)Z_e/f_y\\
Z_e
\end{bmatrix},
\label{eq:center_candidates_3d}
\end{equation}
\begin{equation}
 a_s^C=\operatorname{normalize}
 \begin{bmatrix}
 -s\sin\tau\,d_{m,x}\\
 -s\sin\tau\,d_{m,y}\\
 \cos\tau
 \end{bmatrix}.
\label{eq:axis_candidates}
\end{equation}
Together, $C_s^C$ and $a_s^C$ define the hypothesis
$\mathcal P_s=(C_s^C,a_s^C)$ for $s\in\{-1,+1\}$.

The ellipse alone does not determine the physically consistent sign. The body
mask supplies its image center $p_{\rm body}^I$ and dominant longitudinal
direction $d_b$. Let $g_s$ be the normalized image projection of $a_s^C$.
Body-axis and body-pointing consistency are
\begin{equation}
\begin{aligned}
q_{\rm axis}(\mathcal P_s)&=\frac{1+|g_s^Td_b|}{2},\\
q_{\rm body}(\mathcal P_s)&=\frac{1}{2}\!\left(1+g_s^T
\frac{p_{\rm body}^I-p_s}{\|p_{\rm body}^I-p_s\|+\epsilon}\right).
\end{aligned}
\label{eq:body_cues}
\end{equation}
A centerline term $q_{\rm center}\in[0,1]$ decreases with the normalized
perpendicular distance from $p_s$ to the estimated body centerline. The
body-based score is
$S_{\rm rgb}=w_bq_{\rm body}+w_cq_{\rm center}+w_aq_{\rm axis}$.
When two reliable side-generator lines are available, their oriented common
direction $d_{\rm side}$ yields
\begin{equation}
S_{\rm side}=\gamma_l\frac{1+g_s^Td_{\rm side}}{2}
+\gamma_t q_{\rm body}.
\label{eq:side_score}
\end{equation}
The side score is used only when the line pair passes gates on residual,
support, axis alignment, parallelism, and separation. Thus,
\begin{equation}
S_{\rm dis}(\mathcal P_s)=
\begin{cases}
S_{\rm side}(\mathcal P_s),&Q_{\rm side}\geq Q_{\min},\\
S_{\rm rgb}(\mathcal P_s),&Q_{\rm side}<Q_{\min}.
\end{cases}
\label{eq:disambiguation}
\end{equation}
where $Q_{\rm side}$ is the line-pair quality score. The selected hypothesis is
$s^*=\arg\max_{s\in\{-1,+1\}}S_{\rm dis}(\mathcal P_s)$. The planner receives
$p_o^C=C_{s^*}^C$ and $a^C=a_{s^*}^C$, with $a^C$ directed from the opening
into the cavity. The two semantic masks play different roles: the opening rim
provides metric scale and plane inclination, whereas the body extent and side
generators resolve the otherwise ambiguous direction along the projected minor
axis. This division prevents noisy body boundaries from directly determining
the aperture geometry while still exploiting them for sign disambiguation.
Near-frontal views remain ill-conditioned because $\rho\rightarrow1$ makes
the minor-axis direction noise-sensitive.

\subsection{Target and Observation Model}
\label{subsec:observation_model}

With cavity length $L$ and inner radius $r_{\rm in}$, the task representation is
$\mathcal M^C=\{p_o^C,a^C,L,r_{\rm in}\}$ and the bottom center is
$p_b^C=p_o^C+La^C$. It is transformed to the manipulator base using the
fixed eye-in-hand transform,
\begin{equation}
p^B={}^BR_Cp^C+{}^Bt_C,\qquad a^B={}^BR_Ca^C.
\label{eq:target_transform}
\end{equation}
For joint vector $q\in\R^n$, forward kinematics gives
\begin{equation}
{}^BT_C(q)={}^BT_E(q){}^ET_C=
\begin{bmatrix}
R_c(q)&p_c(q)\\
0&1
\end{bmatrix},
\label{eq:camera_fk}
\end{equation}
with optical axis $z_c(q)=R_c(q)[0,0,1]^T$. A camera-side consistency check
reverses $a$ when necessary so that the current camera lies outside the
opening, then updates $p_b=p_o+La$.

The observation geometry is summarized by
\begin{equation}
\begin{aligned}
e_{\rm axis}(q)&=\arccos\!\left(\clip(z_c(q)^Ta,-1,1)\right),\\
d_{\rm axial}(q)&=(p_c(q)-p_o)^T(-a),\\
e_{\rm lat}(q)&=\left\|p_c(q)-p_o-d_{\rm axial}(q)(-a)\right\|.
\end{aligned}
\label{eq:observation_geometry}
\end{equation}
The valid observation region is
\begin{equation}
\begin{aligned}
\mathcal C_{\rm obs}=\{q~|~&e_{\rm axis}(q)\leq\theta_{\max},\;
 e_{\rm lat}(q)\leq e_{\rm lat,max},\\
&d_{\min}\leq d_{\rm axial}(q)\leq d_{\max}\}.
\end{aligned}
\label{eq:observation_constraints}
\end{equation}
The three terms have distinct geometric roles. $e_{\rm axis}$ controls the
viewing direction, $e_{\rm lat}$ measures the camera's perpendicular distance
from the cavity centerline, and $d_{\rm axial}$ places the camera within an
allowable interval outside the opening. Rotation about the optical axis remains
unconstrained by the inspection task and can therefore be selected according to
robot motion and joint-limit preferences. This retained freedom is precisely
what is lost when one complete Cartesian camera pose is fixed in advance.

To encourage the optical axis to pass through the opening toward the bottom,
a virtual boresight point and gaze error are
\begin{equation}
\begin{aligned}
p_{\rm proxy}&=p_o+(\alpha L+d_{\rm proxy})a,\\
g(q)&=\frac{p_{\rm proxy}-p_c(q)}{\|p_{\rm proxy}-p_c(q)\|},\\
e_{\rm gaze}(q)&=\arccos\!\left(\clip(z_c(q)^Tg(q),-1,1)\right).
\end{aligned}
\label{eq:gaze_error}
\end{equation}
Unlike Eq.~\eqref{eq:observation_constraints}, this term is a soft preference.

\subsection{Feasibility-First Joint-Space Search}
\label{subsec:optimizer}

The method searches joint space directly because the observation region does
not identify a unique camera pose. Multiple seeds are sampled around the current
configuration $q_0$. From each seed, a derivative-free coordinate direct
search evaluates $q\pm\Delta e_i$ and contracts $\Delta$ after an unsuccessful
coordinate sweep \cite{kolda2003directsearch}. Candidates are compared
lexicographically by hard feasibility and then soft quality.

Let $[x]_+=\max(0,x)$ and $r_i=q_{i,\max}-q_{i,\min}$. Define the
normalized lower- and upper-bound violations as
$v_i^-=[q_{i,\min}-q_i]_+/r_i$ and $v_i^+=[q_i-q_{i,\max}]_+/r_i$.
Their aggregate cost is
\begin{equation}
E_{\rm bound}(q)=\sum_i\left[(v_i^-)^2+(v_i^+)^2\right].
\label{eq:bound_cost}
\end{equation}
The hard cost is
\begin{equation}
\begin{aligned}
J_{\rm hard}(q)={}&w_a^h[e_{\rm axis}-\theta_{\max}]_+^2
+w_l^h[e_{\rm lat}-e_{\rm lat,max}]_+^2\\
&+w_s^h[d_{\min}-d_{\rm axial}]_+^2
+w_s^h[d_{\rm axial}-d_{\max}]_+^2\\
&+w_b^hE_{\rm bound}(q).
\end{aligned}
\label{eq:hard_cost}
\end{equation}
For feasible states, let
$\delta_i=\operatorname{wrap}(q_i-q_{0,i})/r_i$,
$E_{\rm motion}=\sum_i\delta_i^2$, and
$E_{\rm wrist}=\sum_{i\in\mathcal W}\delta_i^2$. The normalized valid margin is
$\bar m_i=\min(q_i-q_{i,\min},q_{i,\max}-q_i)/r_i$, giving
\begin{equation}
E_{\rm limit}(q)=\sum_i[\bar m_0-\bar m_i(q)]_+^2.
\label{eq:limit_margin}
\end{equation}
The quality objective is
\begin{equation}
\begin{aligned}
J_{\rm quality}(q)={}&w_mE_{\rm motion}+w_wE_{\rm wrist}+w_jE_{\rm limit}\\
&+w_a^qe_{\rm axis}^2+w_l^qe_{\rm lat}^2
+w_s^q(d_{\rm axial}-d_{\rm desired})^2\\
&+w_ge_{\rm gaze}^2.
\end{aligned}
\label{eq:quality_cost}
\end{equation}
The nominal solution is the lexicographic minimizer of
$(J_{\rm hard},J_{\rm quality})$. Rather than returning only this state, the
solver retains the $N$ highest-ranked distinct feasible configurations,
\begin{equation}
\mathcal Q_{\rm cand}=\{q^{(1)},\ldots,q^{(N)}\},\qquad q^{(1)}=q^*.
\label{eq:candidate_set}
\end{equation}
The lexicographic comparison prevents a reduction in motion cost from
compensating for a violated observation constraint. Retaining multiple distinct
feasible states also separates two forms of feasibility: the optimizer checks
viewing geometry and joint limits, whereas the downstream planner checks
collision-free connectivity from the current state. A highly ranked candidate
can therefore be rejected without rerunning perception or committing the system
to one alternative view.

\subsection{Execution and Post-Execution Verification}
\label{subsec:verification}

MoveIt evaluates $\mathcal Q_{\rm cand}$ in rank order. A candidate is executed
only if a collision-free trajectory can be planned from the current state; a
planning failure advances to the next candidate. The measured robot state then
provides the achieved camera pose, from which the observation errors are
recomputed.

Interior visibility is estimated by sampling points $p_i$ on the circular
bottom. The ray from camera center $p_c$ to $p_i$ intersects the opening plane at
\begin{equation}
\lambda_i=\frac{a^T(p_o-p_c)}{a^T(p_i-p_c)},\qquad
x_i=p_c+\lambda_i(p_i-p_c).
\label{eq:ray_intersection}
\end{equation}
The sample is visible when $0<\lambda_i<1$ and
\begin{equation}
\left\|(I_3-aa^T)(x_i-p_o)\right\|\leq r_{\rm in}-r_{\rm clear},
\label{eq:aperture_test}
\end{equation}
where $I_3$ is the $3\times3$ identity matrix.
The bottom-visible ratio is $V=N_{\rm visible}/N_{\rm total}$. An inspection
succeeds only if the achieved pose remains in $\mathcal C_{\rm obs}$ and
$V\geq V_{\min}$. Because these quantities are recomputed from the measured
post-execution joint state, the acceptance test accounts for discrepancies
between the nominal optimizer output and the configuration actually reached.
The aperture test also captures the finite rim explicitly: satisfying only an
axis-angle condition does not guarantee that rays to the cavity bottom pass
through the opening.

%% file: fig/method_pipeline.tex
\begin{tikzpicture}[
    >=Latex,
    node distance=5.0mm,
    stage/.style={draw, rounded corners=1.2mm, align=center, minimum height=16mm,
                  text width=39mm, inner sep=2.3mm, fill=gray!7},
    flow/.style={->, line width=0.6pt},
    lab/.style={font=\scriptsize, align=center}
]
\node[stage] (p) {\textbf{Target perception}\\
\scriptsize body/rim masks $\rightarrow$ ellipse\\
opening center $p_o^C$, directed axis $a^C$};
\node[stage, right=of p] (o) {\textbf{Observation model}\\
\scriptsize base-frame target geometry\\
axis, lateral, and standoff constraints};
\node[stage, right=of o] (j) {\textbf{Joint-space search}\\
\scriptsize multistart coordinate search\\
feasibility first; rank $\mathcal Q_{\rm cand}$};
\node[stage, right=of j] (e) {\textbf{Execution and verification}\\
\scriptsize MoveIt trajectory planning\\
achieved geometry and bottom visibility};
\draw[flow] (p) -- node[lab,above] {task semantics} (o);
\draw[flow] (o) -- node[lab,above] {$\mathcal C_{\rm obs}$} (j);
\draw[flow] (j) -- node[lab,above] {ranked goals} (e);
\end{tikzpicture}

%% file: sections/04_experiments.tex
\section{Experiments}
\label{sec:experiments}

The evaluation addresses three questions. First, when every method receives the
same target geometry, does observation-constrained joint-space search improve
end-to-end inspection success? Second, do the resulting viewpoints expose more
of the cavity bottom? Third, can the full RGB perception-to-execution pipeline
operate on the tabletop system and on the Unitree A2-mounted manipulator?

\subsection{Setup and Protocol}
\label{subsec:setup}

The platform uses a 6-DoF AgileX Piper manipulator and an eye-in-hand Intel
RealSense D435i. Isaac Sim provides the simulated robot and target, while MoveIt
performs collision-aware planning in simulation and on hardware. The target has
inner diameter 5~cm, outer diameter 6~cm, and length 5.5~cm. The same kinematic
chain, joint limits, candidate-to-MoveIt interface, and post-execution
acceptance test are used throughout the planner comparison.

The proposed optimizer uses 48 seeds within 0.35~rad of $q_0$, at most 300
coordinate updates per seed, an initial step of 0.14~rad, and a halving schedule
down to $5\times10^{-4}$~rad. It retains at most 16 feasible configurations.
The observation thresholds are $\theta_{\max}=0.15$~rad,
$e_{\rm lat,max}=0.005$~m, and $d_{\rm axial}\in[0.08,0.38]$~m, with
$d_{\rm desired}=0.20$~m, $r_{\rm clear}=0.002$~m, and $V_{\min}=0.85$.
These values are fixed for all 100 test configurations rather than retuned for
individual target poses.

Simulation evaluates 100 target configurations with target depth from 0.40 to
0.70~m and axis tilt up to approximately $50^\circ$. Every method receives the
same target configurations, initial robot states, and simulator ground-truth
opening center and axis. This protocol intentionally removes perception error
and isolates viewpoint generation. Only the viewpoint-generation strategy
changes; collision-aware planning, execution, and the final geometric and
visibility criteria remain identical. Success is end-to-end: at least one
candidate must admit a MoveIt trajectory, and the achieved pose must satisfy all
observation constraints and the visibility threshold.

\begin{figure}[t]
    \centering
    \begin{minipage}[b]{0.49\columnwidth}
        \centering
        \includegraphics[width=\linewidth]{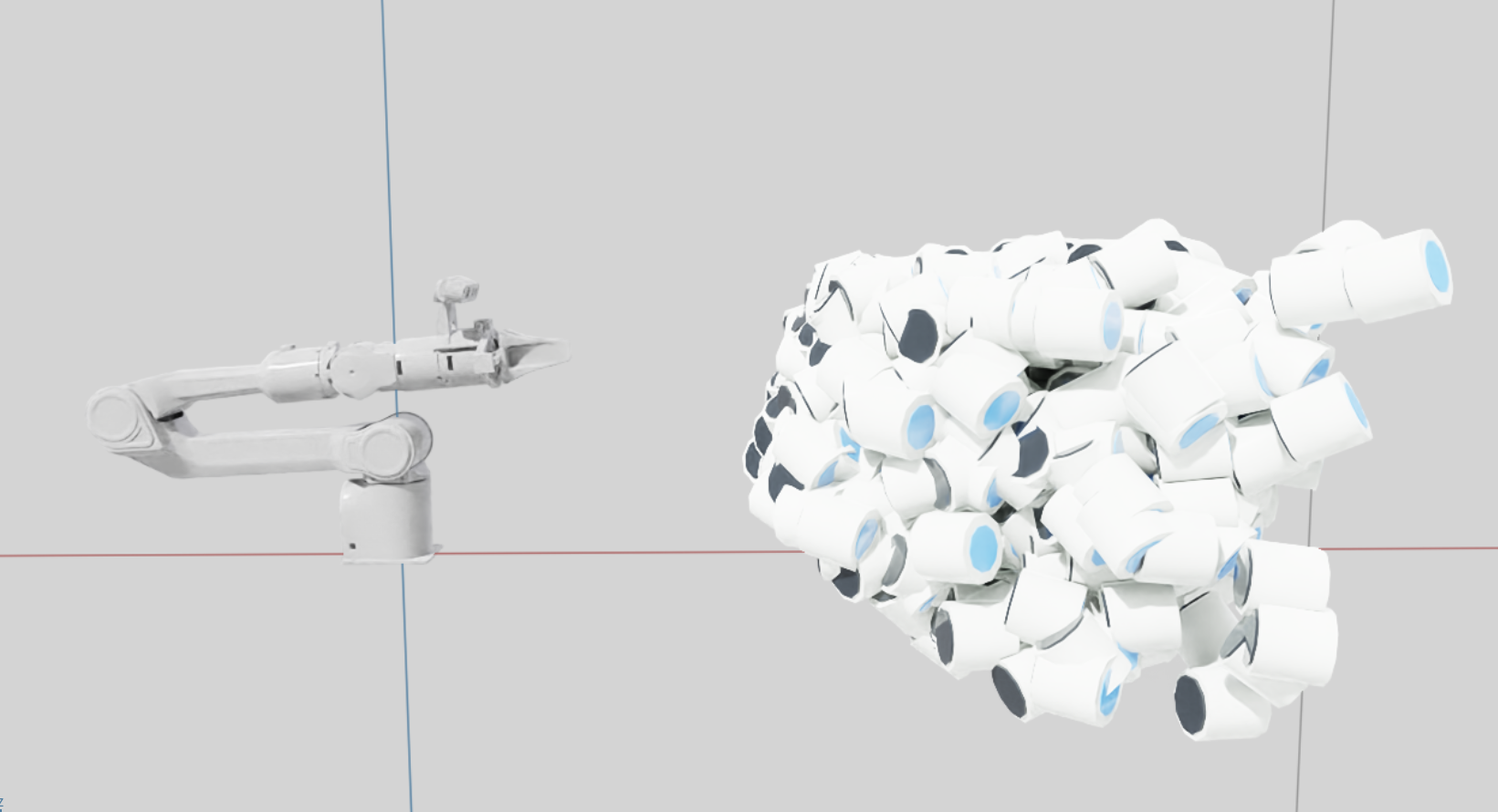}\\[-0.5mm]
        \footnotesize (a)
    \end{minipage}\hfill
    \begin{minipage}[b]{0.49\columnwidth}
        \centering
        \includegraphics[width=\linewidth]{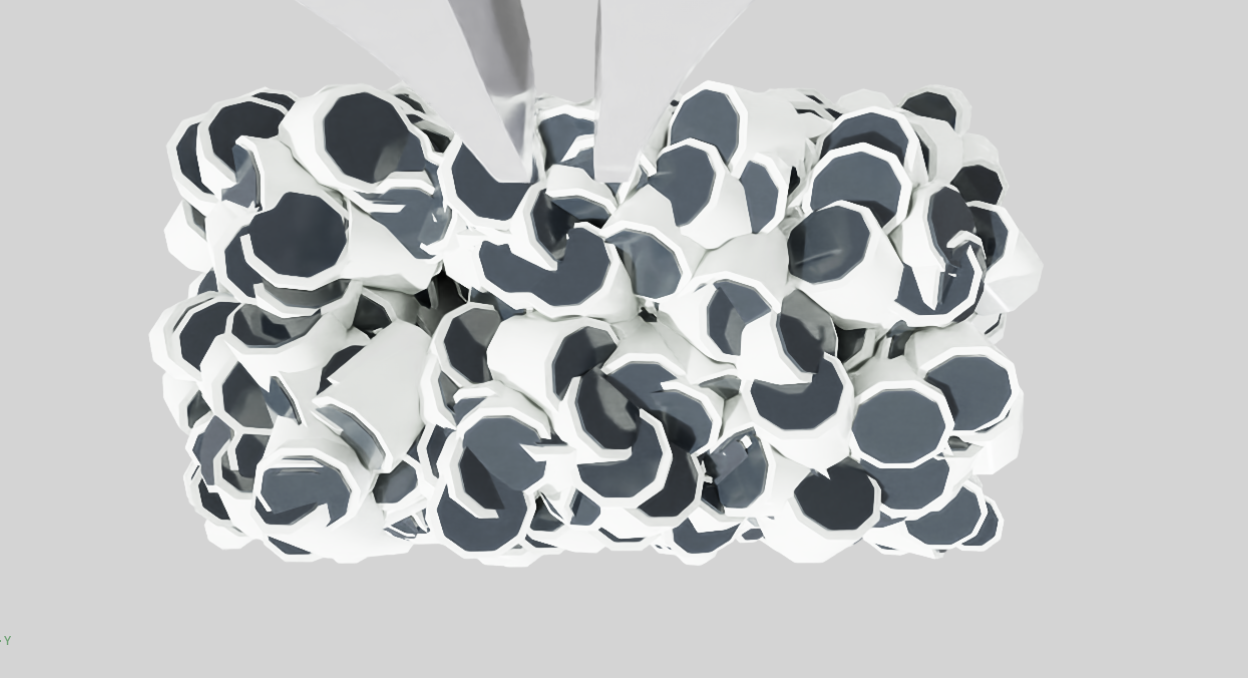}\\[-0.5mm]
        \footnotesize (b)
    \end{minipage}
    \caption{Isaac Sim evaluation visualization: (a) the manipulator and the
    sampled cylindrical-target configurations; (b) a representative
    eye-in-hand RGB observation. The RGB rendering is used for qualitative
    visualization in the controlled comparison, while all methods receive the
    same simulator ground-truth opening center and axis.\vspace{-0.5cm}}
    \label{fig:isaac_simulation}
\end{figure}

\subsection{Simulation Comparison}
\label{subsec:simulation}

Four strategies are compared. \emph{Fixed Pose + IK} constructs one Cartesian
view from the target geometry before solving IK. \emph{View Sampling + IK}
samples multiple Cartesian views and tests their IK solutions.
\emph{Multi-Start CD} applies multistart coordinate search to a conventional
single-view objective. \emph{Proposed} uses the observation region, feasibility
priority, ranked candidate set, and post-execution criteria in
Sec.~\ref{sec:method}. The first two baselines therefore commit to pose-space
choices before manipulator feasibility is known, whereas both coordinate-search
methods operate in joint space. The distinction between the latter two is the
set-valued observation formulation and lexicographic feasibility treatment.

The evaluation reports position error, angular error, bottom-visible ratio, and
inspection success. Success is computed over all 100 configurations. The three
continuous quantities are averaged over trials in which a trajectory is
executed and an actual camera pose is available, including executed trials that
fail the final inspection criterion. A configuration for which every candidate
fails motion planning is counted as an inspection failure but cannot contribute
to an executed-pose average. Because inspection admits multiple valid views,
the pose-error quantities are treated as diagnostic components; the primary
criterion is whether the achieved pose jointly satisfies the observation and
visibility requirements.

\begin{table}[t]
\centering
\caption{Simulation results over 100 target configurations. Position error,
angular error, and visibility are averaged over trials with an executed camera
pose; success is computed over all 100 configurations.}
\label{tab:simulation}
\small
\setlength{\tabcolsep}{2.15pt}
\begin{tabular}{@{}lrrrr@{}}
\toprule
\textbf{Method} & \textbf{Pos.} & \textbf{Ang.} & \textbf{Visible} & \textbf{Success}\\
& \textbf{(mm)} & \textbf{(deg)} & \textbf{(\%)} & \textbf{(/100)}\\
\midrule
Fixed Pose + IK   & 140.9 & 33.60 & 44.58 & 34\\
View Sampling + IK& 49.2  & \textbf{15.51} & 80.2  & 65\\
Multi-Start CD    & \textbf{37.9} & 28.41 & 84.3  & 76\\
Proposed          & 42.9  & 21.68 & \textbf{91.65} & \textbf{92}\\
\bottomrule
\end{tabular}
\end{table}

Table~\ref{tab:simulation} shows that the proposed method succeeds in 92
configurations, improving absolute success by 16 percentage points over
Multi-Start CD and by 27 points over View Sampling + IK. Mean bottom visibility
also rises from 84.3\% for Multi-Start CD to 91.65\%. The lowest position and
angular errors are achieved by different baselines, while neither baseline
attains the highest end-to-end success. This result is consistent with the task
formulation: minimizing one deviation from a nominal view does not ensure that
the executed camera simultaneously remains inside the axial and lateral bounds,
faces through the opening, and exposes enough of the finite bottom surface.

The non-success cases include configurations for which the search does not find
a feasible joint state and cases where a feasible state cannot be connected by
the motion planner. The current experiment records the final end-to-end outcome
but does not retain a complete stage-wise failure breakdown. Consequently, the
results support the benefit of the combined viewpoint-generation interface, but
they do not isolate how much of the remaining 8\% failure rate originates in
search coverage versus trajectory connectivity.

\subsection{Hardware Demonstration}
\label{subsec:hardware}

The full RGB-to-execution pipeline was first run with the Piper fixed to a table
and was then transferred without changing the planner to the Piper mounted on a
Unitree A2. Unlike the controlled simulation comparison, hardware uses the
estimated $p_o^C$ and $a^C$ from Sec.~\ref{subsec:perception}. The tabletop setup
in Fig.~\ref{fig:system_overview}(a) permits repeatable placement around the
multi-opening test fixture, while the A2-mounted setup in
Fig.~\ref{fig:system_overview}(b) evaluates the same eye-in-hand workflow on an
integrated mobile inspection platform.

\begin{figure}[t]
    \centering
    \includegraphics[width=\columnwidth]{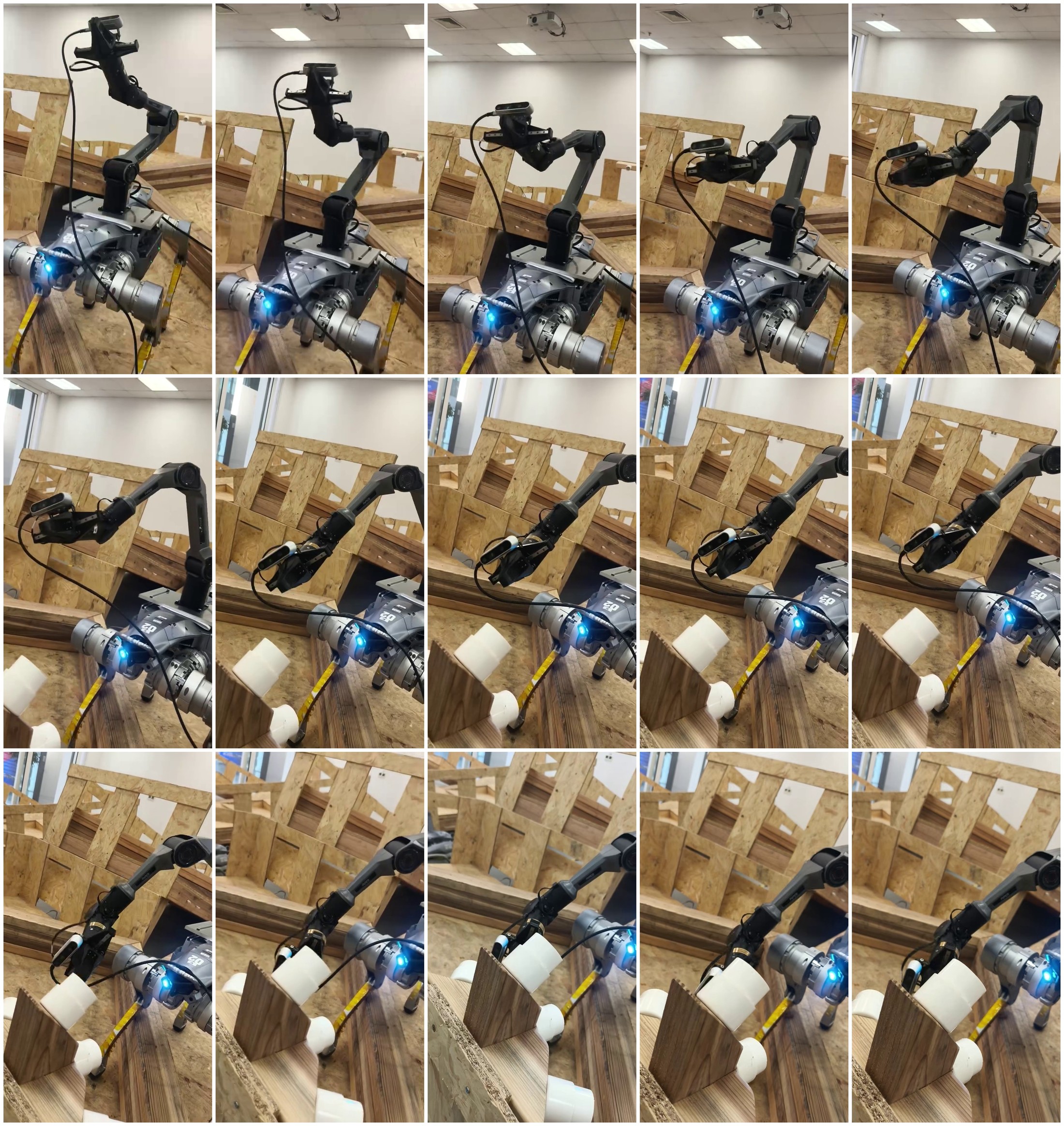}
    \caption{Complete A2-mounted inspection sequence, ordered left-to-right and
    top-to-bottom. The arm leaves its initial posture, reorients toward the
    detected cylindrical opening, approaches the target, and reaches a camera
    configuration from which the cavity interior is visible.\vspace{-0.5cm}}
    \label{fig:hardware_sequence}
\end{figure}

Fig.~\ref{fig:hardware_sequence} illustrates the complete motion rather than
only its initial and final states. The first row shows the large-scale arm
reconfiguration, the second row shows approach to the target fixture, and the
third row shows final alignment near the opening. The sequence demonstrates that
the ranked-viewpoint interface can be executed on the quadruped-mounted arm and
that the final eye-in-hand view exposes the cylindrical interior.

\subsection{Scope and Limitations}
\label{subsec:limitations}

The simulation experiment intentionally uses ground-truth target geometry;
therefore, the 92\% result characterizes viewpoint generation, motion planning,
and executed-pose verification rather than RGB perception accuracy. The
hardware evidence establishes system feasibility but is not a statistical
success-rate study. Repeated trials, perception error against measured target
geometry, runtime distributions, and stage-wise failure counts are needed for a
complete robustness assessment.

The formulation also assumes known cavity dimensions, a static visible target,
and fixed camera-to-end-effector calibration. Near-frontal views remain weakly
conditioned for monocular axis recovery, and coordinate search may become
costly as the number of seeds or manipulator degrees of freedom increases.
These limitations motivate future work on uncertainty-aware target geometry,
online visual correction, and data-driven seed proposals, while retaining the
explicit geometric acceptance test.

%% file: sections/05_conclusion.tex
\section{Conclusion}
\label{sec:conclusion}

This paper formulated cylindrical-cavity inspection as an observation region
rather than a unique 6-DoF camera pose. A compact RGB front end estimates the
opening center and directed interior axis; explicit axis, lateral, and standoff
constraints then guide a feasibility-first joint-space search. Ranked
alternatives allow collision-aware planning to reject an inaccessible goal,
while post-execution geometry and bottom visibility prevent acceptance based
only on a nominal solution.

Across 100 controlled Isaac Sim configurations, the method achieved 92 of 100
end-to-end successes and 91.65\% mean bottom visibility over executed trials.
The baselines with the smallest position or angular error did not produce the
highest inspection success, supporting the use of a set-valued task criterion.
Tabletop and Unitree A2-mounted demonstrations establish end-to-end feasibility;
quantitative real-world perception and repeated-trial evaluation remain future
work.

Future work will investigate uncertainty-aware constraints, faster seed
selection, and closed-loop image correction. Low-cost VR teleoperation offers a
practical route for collecting demonstrations \cite{dong2025enabling} from
which seed distributions or recovery motions can be learned without replacing
the geometric verification stage.